\documentclass[letterpaper, 10 pt, conference]{ieeeconf}  
\pdfoutput=1
\IEEEoverridecommandlockouts
\usepackage{cite}
\usepackage{amsmath,amssymb,amsfonts}
\usepackage{algorithmic}
\usepackage{graphicx}
\usepackage[hyphens,spaces,obeyspaces]{url}
\usepackage{hyperref}
\usepackage{todonotes}
\usepackage{textcomp}
\usepackage{xcolor}
\usepackage{bm}
\usepackage{threeparttable}
\usepackage{siunitx}
\usepackage{multirow}
\usepackage{comment}
\usepackage{booktabs}
\usepackage{tikz}
\usepackage{subcaption}
\usepackage{hypcap} 
\usepackage{comment}
\usepackage{gensymb}

\usepackage[skip=8pt,font=small]{caption}

\def\BibTeX{{\rm B\kern-.05em{\sc i\kern-.025em b}\kern-.08em
    T\kern-.1667em\lower.7ex\hbox{E}\kern-.125emX}}

\hypersetup{
	colorlinks=true,
	linkcolor=black,
	filecolor=magenta,      
	urlcolor=black,
}

\begin{document}


\title{\LARGE \bf Human Gaze and Head Rotation during Navigation, Exploration and Object Manipulation in Shared Environments with Robots}

\author{Tim~Schreiter$^{3,1}$,
       Andrey Rudenko$^2$,
       Martin Magnusson$^1$,
       and~Achim J.~Lilienthal$^{3,1}$
\thanks{$^{1}$Centre for Applied Autonomous Sensor Systems (AASS),
	\"Orebro University, Sweden { \{tim.schreiter,martin.magnusson\}@oru.se}}
\thanks{$^{2}$Robert Bosch GmbH, Corporate Research, Stuttgart, Germany
{\tt\small andrey.rudenko@de.bosch.com}}%
\thanks{$^{3}$TU Munich, Germany 
{\tt\small achim.j.lilienthal@tum.de}}%
\thanks{This work was supported by the European Union’s Horizon 2020 research and innovation program under grant agreement No. 101017274 (DARKO)}}

\maketitle

\begin{abstract}
The human gaze is an important cue to signal intention, attention, distraction, and the regions of interest in the immediate surroundings. Gaze tracking can transform how robots perceive, understand, and react to people, enabling new modes of robot control, interaction, and collaboration. In this paper, we use gaze tracking data from a rich dataset of human motion (THÖR-MAGNI) to investigate the coordination between gaze direction and head rotation of humans engaged in various indoor activities involving navigation, interaction with objects, and collaboration with a mobile robot.
In particular, we study the spread and central bias of fixations in diverse activities and examine the correlation between gaze direction and head rotation. We introduce various human motion metrics to enhance the understanding of gaze behavior in dynamic interactions. Finally, we apply semantic object labeling to decompose the gaze distribution into activity-relevant regions.
\end{abstract}

\section{Introduction}\label{sec:intro}

Robots operating in shared environments with humans can benefit significantly from the ability to track and interpret various cues related to human motion and activity. The context of human motion includes a wide range of cues, such as
full-body poses, gestures, gazes, motion velocity, acceleration, and many others.
The ability of robots to interpret these cues is essential for several reasons: enhanced safety by predicting human actions, improved efficiency by anticipating human needs, and promotion of more natural interaction between humans and robots by responding to nonverbal signals.

Gaze has been described as a window into the human mind. It provides information related to human attention and intention. Integrating gaze tracking into Human-Robot Interaction (HRI) approaches can help robots better understand human behavior and, in turn, help robots navigate shared spaces more effectively and participate in collaborative tasks with greater awareness and adaptability.
As such, studying the human gaze in human-robot interaction helps to create robotic systems that smoothly share our spaces and comprehend and anticipate our actions and intentions.

\begin{figure}[t]
    \centering
    \includegraphics[width=0.93\linewidth]{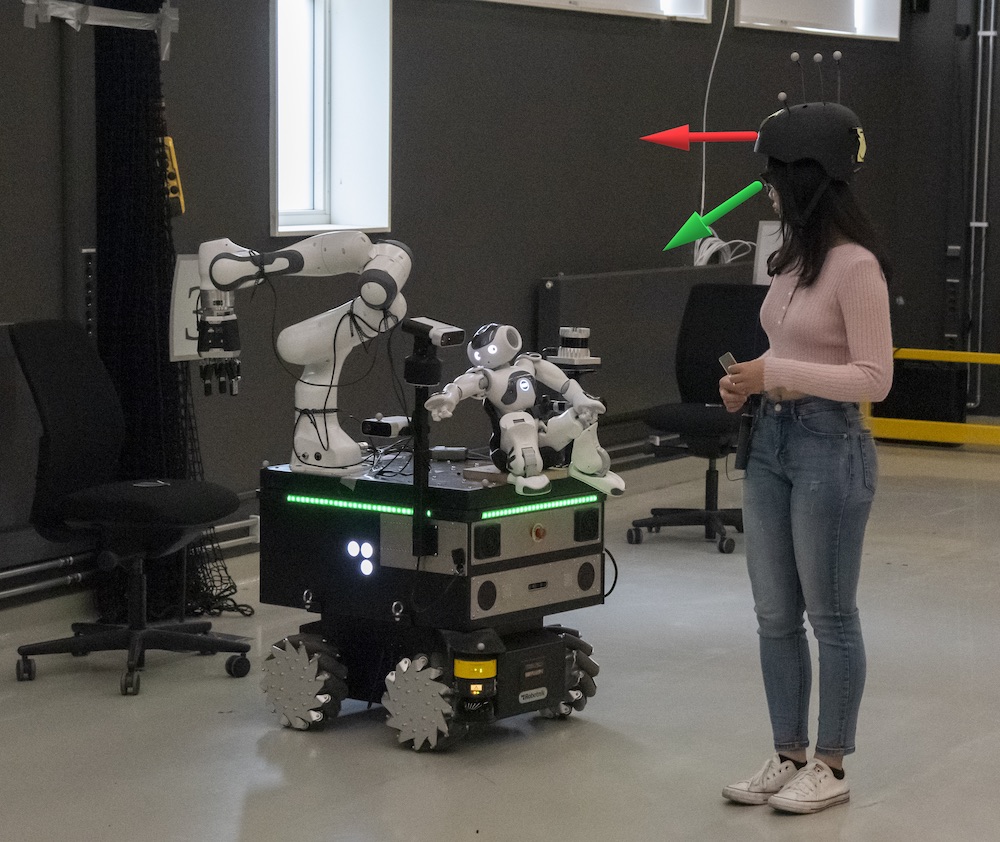} \\ \vspace{3pt}
    \includegraphics[width=0.3\linewidth,height=4cm,keepaspectratio]{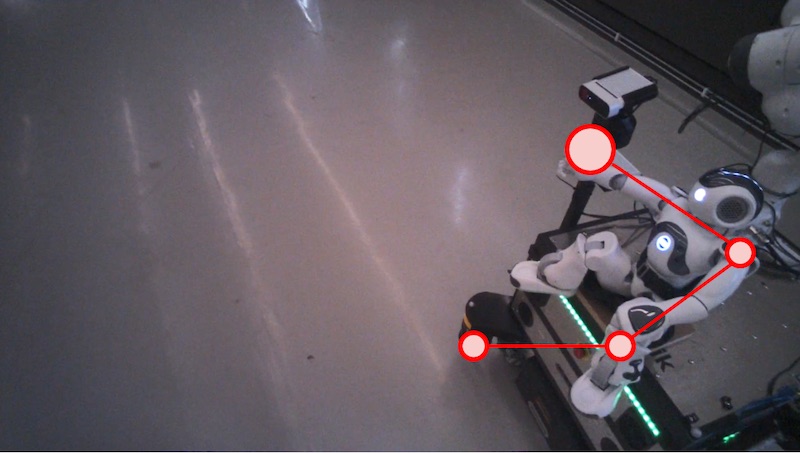}
    \includegraphics[width=0.3\linewidth,height=4cm,keepaspectratio]{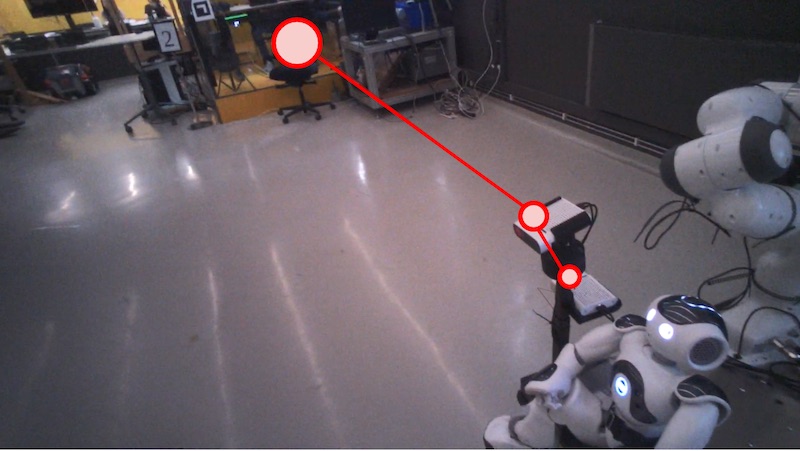}
    \includegraphics[width=0.3\linewidth,height=4cm,keepaspectratio]{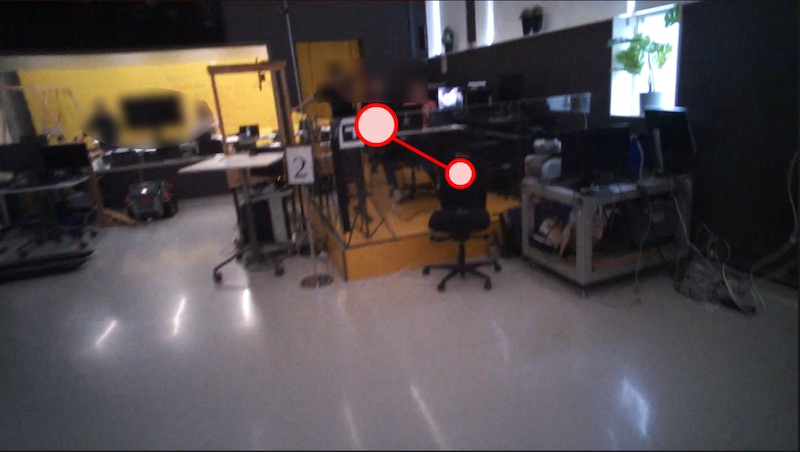}
    \caption{A participant of the THÖR-MAGNI dataset attends to instructions of the mobile robot \cite{schreiter2023advantages}. \textbf{Top:} Illustration of the visual difference between the head orientation (\textbf{red}) and gaze direction (\textbf{green}).
    \textbf{Bottom:} a sequence of gazes on the mobile robot, followed by a shift of attention to the goal point that the robot cued. This shift is followed by a head rotation to center the visual field on the goal point. Fixations are shown with \textbf{white circles}, and their sequences are connected by \textbf{red lines}.}
    \label{fig:cover}
\end{figure}

That being said, studies of human gaze in motion and dynamic human-robot interactions are still scarce, not least due to the complexity of tracking the gaze of a moving person. Thus, head orientation is often used as a proxy for gaze direction and using head orientation has been shown to improve the interaction between humans and robots \cite{dermy2019multi}. Furthermore, head orientation is successfully used in automated driving settings to infer the attention and intention of pedestrians and cyclists \cite{flohr2018vulnerable}.
However, relying solely on head orientation as an indicator of gaze has limitations due to the complex nature of human attention, which often involves subtle eye movements not captured by head orientation alone \cite{palinko2016robot}, (see Figure~\ref{fig:cover}). 

This study analyzes the gaze patterns of people moving and interacting in a dynamic environment shared with robots. We utilize the THÖR-MAGNI dataset, unique for its synchronized data on head orientation, eye movement patterns, and walking trajectories across a diverse group of individuals \cite{Schreiter2024THORMAGNIAL}.
In particular, we show the potential and limitations of using head orientation as a proxy for gaze and the complex relationship between head movements and gaze direction. 

Our study employs various analytical approaches to examine and describe human gaze patterns. Firstly, we focus on the distribution of visual fixations on the 2D tracker plane to evaluate the uncertainty caused by eye rotation relative to head orientation. We extend the analysis of fixations by examining participants' activities and the specific micro-actions they performed during tasks and interactions. We use heatmaps to visualize fixations and identify patterns of visual engagement and attention allocation.

To offer a geometric representation of where participants fixated most frequently on these heatmaps in the 2D tracker plane, we apply ellipse-fitting techniques to summarize and analyze areas of highest fixation density, referred to as "central tendencies." Additionally, the levels of engagement are quantified by calculating the average duration and rate of fixations. This allows for a deeper understanding of how participants interacted with their environment and the robots within it. Through this analysis, we aim to provide more effective support for gaze-informed predictions in dynamic settings and highlight the nuanced ways human attention is directed and sustained during human-robot interaction.

Furthermore, we investigate the coordination between eye and head movements during attention shifts. We compare our findings in the indoor settings with prior studies in outdoor environments. We correlate head orientation and gaze vectors with motion metrics to link visual attention with physical movement. With this analysis, we seek to support the deployment of appearance-based gaze estimation methods, which struggle with head and eye coordination variability\cite{ghosh2023automatic}, especially in dynamic environments.

Lastly, we leverage the YOLO object detection model to qualify the objects human gaze at more precisely. By identifying and categorizing objects or areas that attract significant visual focus, we gain insights into the semantics of targets of participants' gaze, enriching our understanding of attention allocation in dynamic settings, especially during locomotion. Applying modern computer vision techniques to eye-tracking data is a promising approach to contextually interpreting human attention within the context of HRI.

The paper is organized as follows: in Sec. \ref{sec:RL}, we review and motivate the use of human gaze in robotics applications. In Sec. \ref{sec:Methods}, we present our tools to analyze the human gaze during motion in shared environments. In Sec. \ref{sec:discuss}, we draw insights from the conducted analysis, and Sec.~\ref{sec:concl} concludes the paper. 

\section{Related Work}\label{sec:RL}

The human gaze plays an increasingly important role in various robotic applications. Gaze tracking has long been used by social robots, for instance, in conversations to manage turn-taking, improve the information exchange, and reinforce mutual understanding. Gaze tracking is useful in collaborative robots, such as handovers, to coordinate the joint maneuver \cite{grigore2013joint}. Gaze can also be a control technique to reference objects \cite{AdmoniReview}. Gaze tracking enables execution of anticipatory control actions \cite{huang2016anticipatory}, in particular in hybrid bionic systems such as exo-skeletons \cite{pinpin2008gaze}, and aids collaborative search tasks \cite{brennan2008coordinating}.

In learning tasks, robots can learn from how humans distribute their attention to other moving people to achieve more efficient and natural crowd navigation \cite{chen2020robot}. Human gaze tracking can help focus robot attention in imitation learning by limiting the sensor input and the number of irrelevant objects and relations processed \cite{kim2020using}. Measuring gaze is useful in behavior recognition \cite{yi2009recognizing} and in driver behavior modeling \cite{liu2001modeling}, for instance, to find correlations between certain fixation patterns and driving tasks, aiming to detect driver behavior and intention. Finally, models and metrics to describe gaze patterns are useful to mimic human behavior in robot gaze applications \cite{AdmoniReview,kshirsagar2020robot}.


A natural correlation exists between gaze direction and head orientation \cite{freedman2008coordination, foulsham2011and, franchak2021adapting}. Researchers used this natural correlation for robotics applications to achieve natural and effective HRI \cite{haefflinger2023benefit}. In autonomous driving applications, the studies of head movements preceding eye movements highlight the potential of coordinated gaze behavior for goal-directed visual scanning and adapting behaviors to environmental changes \cite{liu2022control}. Our study explores the contribution of eyes and the head to shifts of attention in various activities and compares these findings to outdoor environments \cite{franchak2021adapting}. 

Head orientation offers a rough estimate of the human gaze. However, eye-tracking proves more fitting in contexts like social interactions within robotics \cite{palinko2016robot, AdmoniReview} or robot manipulation tasks \cite{kim2020using} due to its sensitivity to subtle cues and ability to filter out irrelevant objects. Eye gaze is crucial in these scenarios, prompting analytical tools like 2D heat maps and areas of interest (AOI) to study attention distribution and eye-tracking utility in Human-Robot Interactions \cite{kompatsiari2019measuring, schreiter2023advantages}. 


Understanding the semantics of human gaze and attention in environments shared by robots and humans is crucial for enhancing robotics applications, especially in dynamic settings with numerous potential distractors. 
In our study, we use the YOLOv8 model \cite{yolov8_ultralytics} with a subset of the THÖR-MAGNI eye-tracking data to investigate its potential to interpret the visual attention when navigating and interacting with people. The findings indicate that human attention towards a mobile robot in a shared environment remains constant, regardless of the participant's activity or navigational behavior. 

\section{Analysis of Gaze Patterns in Navigation and Interaction Tasks}\label{sec:Methods}
\label{sec:magni}

In this paper, we develop and describe a methodology to analyze human gaze in dynamic environments. We study and compare gaze behavior across various activities and tasks, which include search and navigation towards goals in the room, manipulation of objects, social interactions and receiving instructions from the robot. We have two goals in mind: first, to provide tools and methods to support human activity understanding and prediction from mobile gaze trackers, and second, to quantify human gaze in relation to head orientation in scenarios where systems may need to rely on the head orientation as a proxy for head direction.

In this section we present the dataset of human gaze and motion used in our analysis (Sec.~\ref{subsec:thor}), followed by the analysis of gaze distribution and its bias in Sec.~\ref{subsec:heatmaps}, \ref{subsec:microactions} and \ref{subsec:central_tendencies}. In Sec.~\ref{subsec:head-eye-coordination}, we introduce head orientation into the analysis and discuss the head- and eye-rotation comfort ranges. In Sec.~\ref{subsec:motion}, we discuss auxiliary motion metrics, and, in Sec.~\ref{subsec:objects}, examine the distribution of gazes towards static and dynamic semantic objects in the environment.

\subsection{THÖR-MAGNI Dataset of Human Gaze And Motion}
\label{subsec:thor}

Our study utilizes the THÖR-MAGNI dataset\footnote{Available at \url{http://thor.oru.se/magni.html}} to analyze human gaze and movement in environments shared with robots \cite{Schreiter2024THORMAGNIAL}. This extensive dataset includes indoor motion capture recordings of human activities, ranging from solo and group navigation to transporting objects and interacting with robots. It encompasses data from 40 participants (21 males and 19 females, mean age $30.18\pm6.73$ years) across 52 four-minute sessions and provides 468 minutes of eye-tracking data from 16 participants engaged in different activities.

For this study, we focus on Scenarios 1-3 data, each highlighting different aspects of human motion, following our prior analysis on these scenarios \cite{de_Almeida_2023_ICCV, schreiter2022magni}. Scenario~1 is the baseline, featuring navigation with and without applied semantic attributes (e.g., floor markings and one-way passages). Scenario 2 introduces additional activities for participants in an environment without semantic attributes. Scenario 3 builds upon Scenario 2 by incorporating varied navigation styles of the robot, including a directional (3A) and an omnidirectional driving style (3B).

The data was collected using three eye-tracking devices, which provided comprehensive details on participants' positions, velocities, and head orientations. Participants are categorized based on their activities: ``Visitors,'' who navigate alone or in small groups, and "Carriers," who transport objects like boxes ("Carrier-Box"), buckets (``Carrier-Bucket''), and poster frames (``Carrier-Large Object''). Our study leverages the dataset to explore the dynamics of head orientation, gaze, and attention allocation during navigation, exploration, and object manipulation in human-robot environments, providing valuable insights into human-robot interaction.

\begin{table}
    \centering
    \vspace{3pt}
  \caption{Amount of eye-tracking data available from the Tobii glasses for various activities in the THÖR-MAGNI dataset.}
  \label{tab:recorded_data}
  \resizebox{\columnwidth}{!}{
  \begin{tabular}{lcc}
    \toprule
    Activity&Recorded minutes&Scenario\\
    \midrule
    Visitors-Alone & 108 & All\\
    Visitors-Group 2 & 124 & All\\
    Visitors-Group 3 & 52 & All\\
    Carrier-Bucket & 32 & 2--3\\
    Carrier-Box & 60 & 2--3\\
    Carrier-Large Object & 92 & 2--3\\
  \toprule
  Total & 468
\end{tabular}
}
\end{table}

\subsection{Overall Gaze Distribution}\label{distro_heatmaps}
\label{subsec:heatmaps}

First, we study gaze points where participants focused their attention, known as fixations \cite{blascheck2017visualization}. An example of fixation sequences obtained with eye-tracking glasses is shown in Figure~\ref{fig:cover} (bottom).
We visualize the accumulation of fixations using heatmaps, also called ``attention maps''. These heatmaps, generated from the THÖR-MAGNI dataset's eye-tracking data, illuminate how participants distribute visual attention within environments shared with robots and other humans. To account for the varying geometries of scene cameras across eye-tracking devices, we applied coordinate transformations, aligning their field of view (FOV). We transformed the data to accommodate the differing FOVs of the Tobii~3 Glasses ($95\degree$ horizontal and $63\degree$ vertical), Tobii~2 Glasses ($82\degree$ horizontal and $52\degree$ vertical), and Pupil~Invisible Glasses ($82\degree$ horizontal and $82\degree$ vertical). 

\begin{figure}[t]
    \centering
    \vspace{3pt}
    \includegraphics[width=1\linewidth]{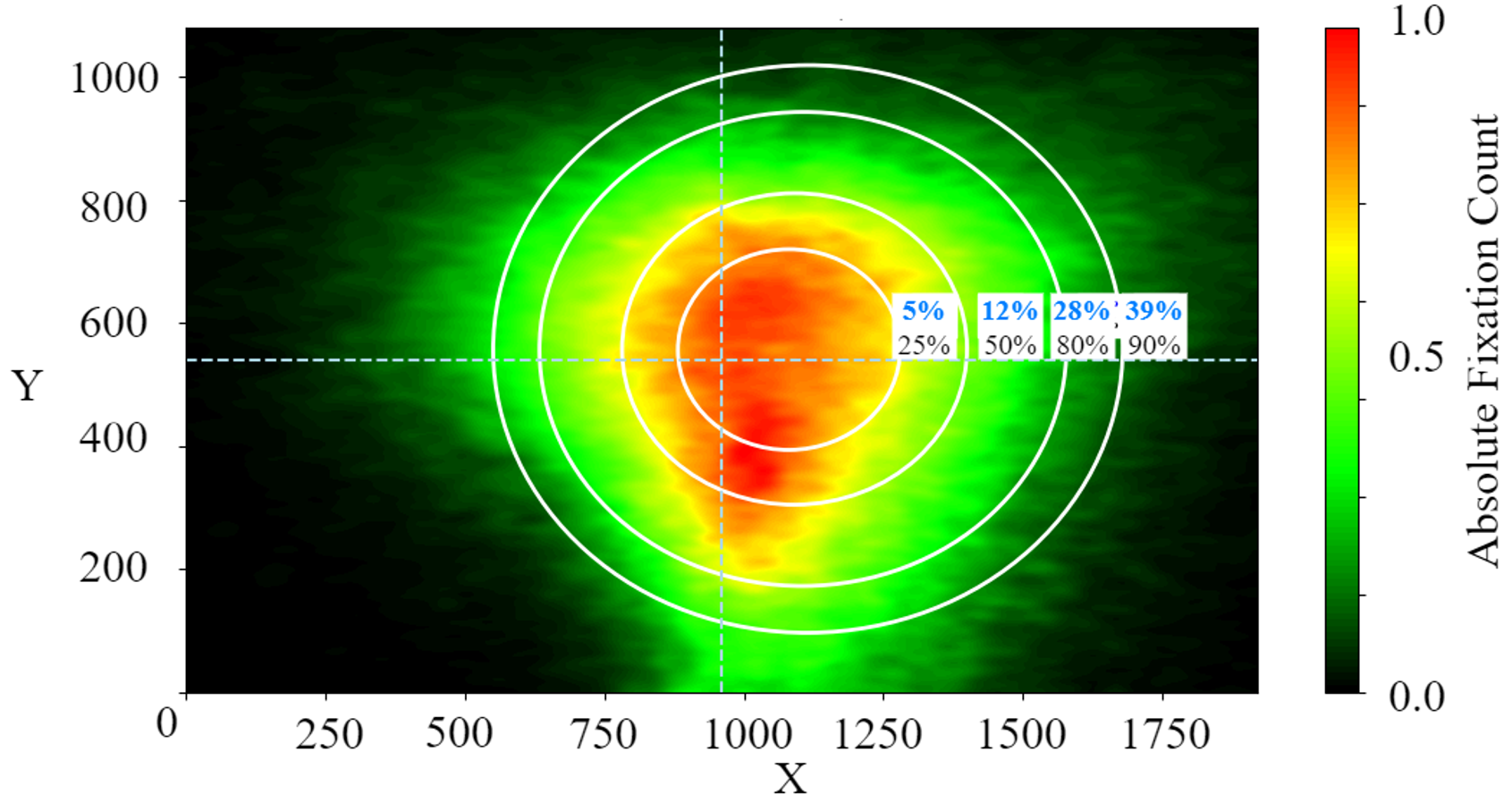}
    \caption{Fixation locations in the THÖR-MAGNI dataset. \textbf{Ellipses} represent areas containing $25\%$, $50\%$, $80\%$, and $90\%$ (\textbf{Black Labels}) of all recorded gazes. \textbf{Blue Labels} indicate the percentage of the 1920x1080 image included in each ellipse.}
    \label{fig:CBS-THOR-MAGNI}
\end{figure}

Figure \ref{fig:CBS-THOR-MAGNI} displays the heatmap created for the entire gaze fixations. We identify a preference for gaze points along the vertical center of the visual field, accompanied by a stronger variation in vertical fixation positions compared to horizontal ones. The participants' focus in our study is slightly shifted to the right of the vertical center line. Additionally, there was a general trend of participants directing their gaze more toward the upper portion of the images.

Using heatmaps, we can also quantify these shifts of the gaze distribution. To that end, we use an ellipse fitting technique from literature \cite{ioannidou2016centrial}, which leverages the covariance matrix, eigenvalues, and chi-squared distributions to accurately delineate areas of concentrated gaze. This method, also visualized in Figure \ref{fig:CBS-THOR-MAGNI}, encapsulates areas representing $25\%$, $50\%$, $80\%$, and $90\%$ of all collected fixations, providing a quantitative measure of where participants' gazes converge most frequently. Our findings reveal that ellipses covering 28\% and 39\% of the image area encapsulate 80\% and 90\% of all fixations, respectively, underscoring participants' central focus in the visual field.

\subsection{Gaze Distribution Across Activities and Micro-Actions}
\label{subsec:microactions}
\begin{figure}[!t]    
\centering
\vspace{5pt}
\includegraphics[width=1\linewidth]{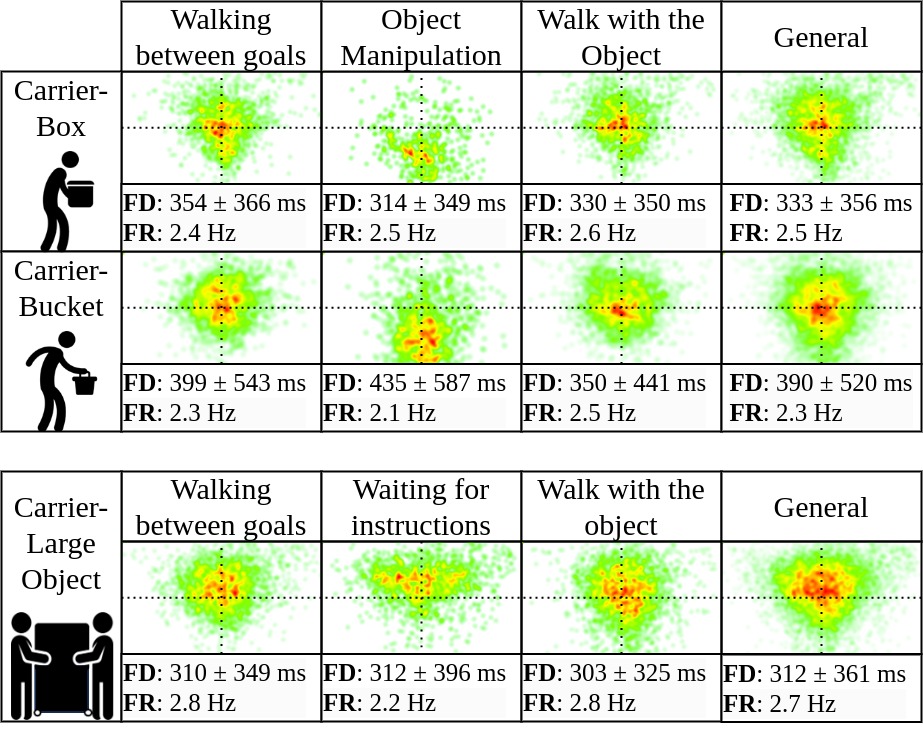}   \includegraphics[width=0.9\linewidth]{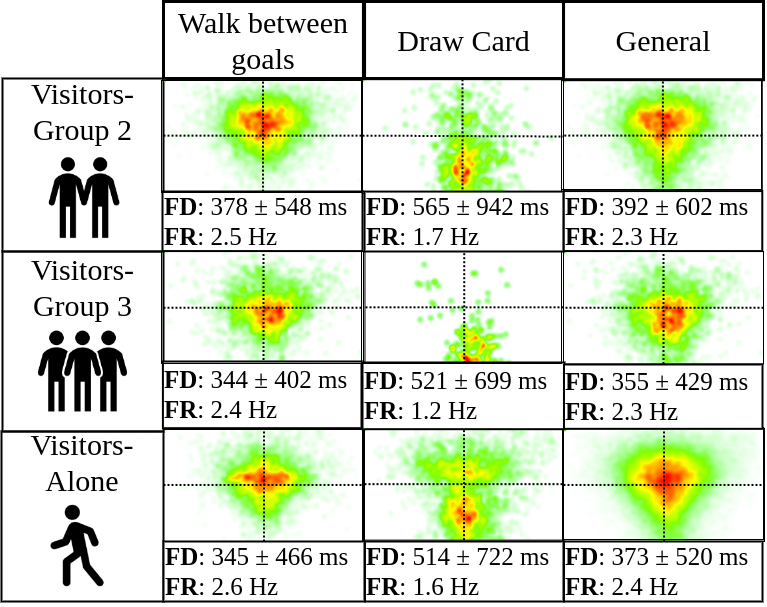}
\caption{Heatmaps of fixation locations in the dataset with average \textbf{Fixation duration (FD)} and overall \textbf{Fixation rate (FR)} per role and the comprised micro-actions. Visual axes are shown with dotted lines. Central coordinates and the spread of the central biases are listed in Tables~\ref{tab:activ-carry}~and~\ref{tab:activ-visit}.}
\label{fig:Heatmaps}
\end{figure}

We examine how gaze distribution varies across participant activities and micro-actions to understand the intricacies of the human gaze in motion and around robots within the THÖR-MAGNI dataset's dynamic settings. We analyze activities such as navigating to goal points and manipulating objects. For consistency reasons, our analysis focuses on the initial three scenarios involving the roles of ``Visitors'' and ``Carriers'' (see \cite{schreiter2022effect} and  \cite{de_Almeida_2023_ICCV}). Therefore, we do not include Scenarios 4 and 5, with a stronger focus on HRI, as there is already a study concerning the human gaze in these scenarios \cite{schreiter2023advantages}. Figure~\ref{fig:Heatmaps} illustrates the spatial distribution of fixations, revealing nuanced visual attention patterns across roles and micro-actions, supported by Tables \ref{tab:activ-carry} and \ref{tab:activ-visit} contain numerical values describing the center of mass and spread of the central fixation tendencies. ``Carrier-Box'' and ``Carrier-Bucket'' participants exhibit concentrated gazes toward the center during object transportation, indicating focused attention necessary for this task. In contrast, the ``Carrier-Large Object'' group shows a more dispersed gaze pattern, particularly favoring the upper hemisphere. This dispersion likely reflects the need for broader environmental awareness in localizing the object and maneuvering oversized items, partially occluding their visual field. ``Visitors-Group 2'' and ``Visitors-Alone'' display varied gaze distributions, highlighting the impact of group size and task complexity. Notably, ``Visitors-Group 3'' participants prefer the lower hemisphere, possibly indicating a different visual engagement strategy due to group dynamics or task demands.

A common thread across activities is a downward gaze bias during tasks requiring direct object interaction, such as drawing cards or manipulating objects. This bias originates from the natural inclination to align one’s gaze directly with the object of interest, which can cause a misalignment between the head and the eye tracker’s scene camera \cite{ioannidou2016centrial}. It highlights the influence of physical interactions on gaze behavior, emphasizing the need to consider these dynamics in designing intuitive human-robot interfaces. Most visitors exhibited a slight rightward gaze tendency towards the upper hemisphere, with greater spread across the horizontal axis during navigation, indicating visual exploration \cite{franchak2021adapting} and significant vertical variability when manipulating cards. Visitors in groups of three included a horizontal component during card manipulation, focusing more on the center and slightly preferring the lower hemisphere. Consistent with the literature, individuals carrying boxes and buckets showed a slight leftward fixation bias and a central focus during object interaction \cite{ioannidou2016centrial}.

\subsection{Quantifying Gaze Distribution: Central Bias and Spread}
\label{subsec:central_tendencies}

\begin{table*}[!]
\centering
\vspace{3pt}
\caption{Central shift of gaze distribution in activities and micro-actions from Figure \ref{fig:Heatmaps} (top). In each cell, the tuple indicates the 2D coordinates of the distribution center of mass with respect to the center of the frame (rounded to the nearest 5 pixels). The percentage indicates the area of the ellipse (that encompasses 80\% of the fixations) with respect to the area of the frame.}
\label{tab:activ-carry}
\resizebox{0.85\textwidth}{!}{
\begin{tabular}{lcccc}
\toprule
\textbf{Activity} & \textbf{Walk between goals} & \textbf{Object Manipulation} & \textbf{Walk with the Object} & \textbf{General} \\
\midrule
Carrier-Box & (-20, 0), 18.6\% & (-40, -240), 10.6\% & (-100, 50), 15.9\% & (0, 20), 27.3\% \\
Carrier-Bucket & (0, 150), 28.5\% & (60, -260), 14.8\% & (-80, 30), 18.2\% & (-10, -40), 19.1\% \\
\addlinespace 
\midrule
& \textbf{Walk between goals} & \textbf{Wait for instructions} & \textbf{Walk with the Object} & \textbf{General} \\
Carrier-Large Object & (0, 75), 19\% & (-35, 150), 22.7\% & (50, 50), 21.2\% & (0, 125), 25\% \\
\bottomrule
\end{tabular}}
\end{table*}

\begin{table*}[!]
\centering
\caption{Central shift of gaze distribution in activities and micro-actions from Figure \ref{fig:Heatmaps} (bottom). In each cell, the tuple indicates the 2D coordinates of the distribution center of mass with respect to the center of the frame (rounded to the nearest 5 pixels). The percentage indicates the area of the ellipse (that encompasses 80\% of the fixations) with respect to the area of the frame.}
\label{tab:activ-visit}
\resizebox{0.62\textwidth}{!}{
\begin{tabular}{lccc}
\toprule
\textbf{Activity} & \textbf{Walk between goals} & \textbf{Draw Card} & \textbf{General} \\
\midrule
Visitors-Group 2 & (100, 50), 22.8\% & (-30, -270), 14.7\% & (10, 250), 24.9\% \\
Visitors-Group 3 & (80, -80), 22.7\% & (80, -420), 8.5\% & (80, -100), 17\% \\
Visitors-Alone  & (80, -100), 15.4\% & (30, -250), 34.1\% & (50, 50), 24.2\% \\
\bottomrule
\end{tabular}}
\end{table*}

This study examines the distribution of visual fixations during various participant activities, including micro-actions such as ``Walk between goals'' or ``Draw Card'' and tasks involving object manipulation. The findings are presented in Tables \ref{tab:activ-carry} and \ref{tab:activ-visit} as tuples indicating the 2D coordinates of the fixations' center of mass relative to the image's central point (illustrated by the crossing of the dotted lines in Figure~\ref{fig:Heatmaps}). Following the methods described in subsection \ref{distro_heatmaps}, we fit ellipses to encompass 80\% of the fixation distributions. We list the percentages that detail the proportion of the 2D eye-tracking plane these occupy alongside the coordinates in the tables. This analysis sheds light on visual attention patterns across different tasks, marking the shifts in focus with precise distances from the image's center. For the ``Carrier-Box'' and ``Carrier-Bucket'' groups, the analysis revealed that the central displacements of their fixation hotspots were generally close to the image's center, indicating a concentrated area of visual attention during most activities, except during ``Object Manipulation,'' where the focal areas significantly diverged from the center. This pattern suggests that tasks requiring detailed object interaction prompt broader visual engagement, as evidenced by the larger displacement values. Conversely, the ``Carrier-Large Object'' group's fixations were predominantly in the image's upper hemisphere, indicating a consistent focus area across their activities.

During the analysis of the visitors' activities, we observe discernible patterns in the distribution of visual attention across different micro-actions. Specifically, during the ``Draw Card'' tasks, there was a noticeable shift of focus toward the lower hemisphere of the image, indicating a heightened level of visual engagement unique to this activity. This gaze concentration is distinct from the more varied attention patterns associated with other tasks, indicating that specific actions can significantly influence where and how visual attention is assigned. In tasks other than ``Draw Card,'' the ``Visitors-Alone'' and ``Visitors-Group 2'' categories exhibited behaviors consistent with active visual exploration and navigation within the space. Conversely, ``Visitors-Group 3'' primarily focused their gaze on the center of the image, indicating a desire to facilitate communication and coordination within the group. These observations highlight the dynamic and task-specific nature of visual attention among the ``Visitors'', emphasizing how the context and demands of different activities subtly shape the collective and individual focus within groups.

\subsection{Quantifying Eye-Head Coordination in Gaze Shifts}
\label{subsec:head-eye-coordination}
\begin{figure}
  \centering
  \begin{subfigure}{\linewidth}
    \includegraphics[width=1\linewidth]{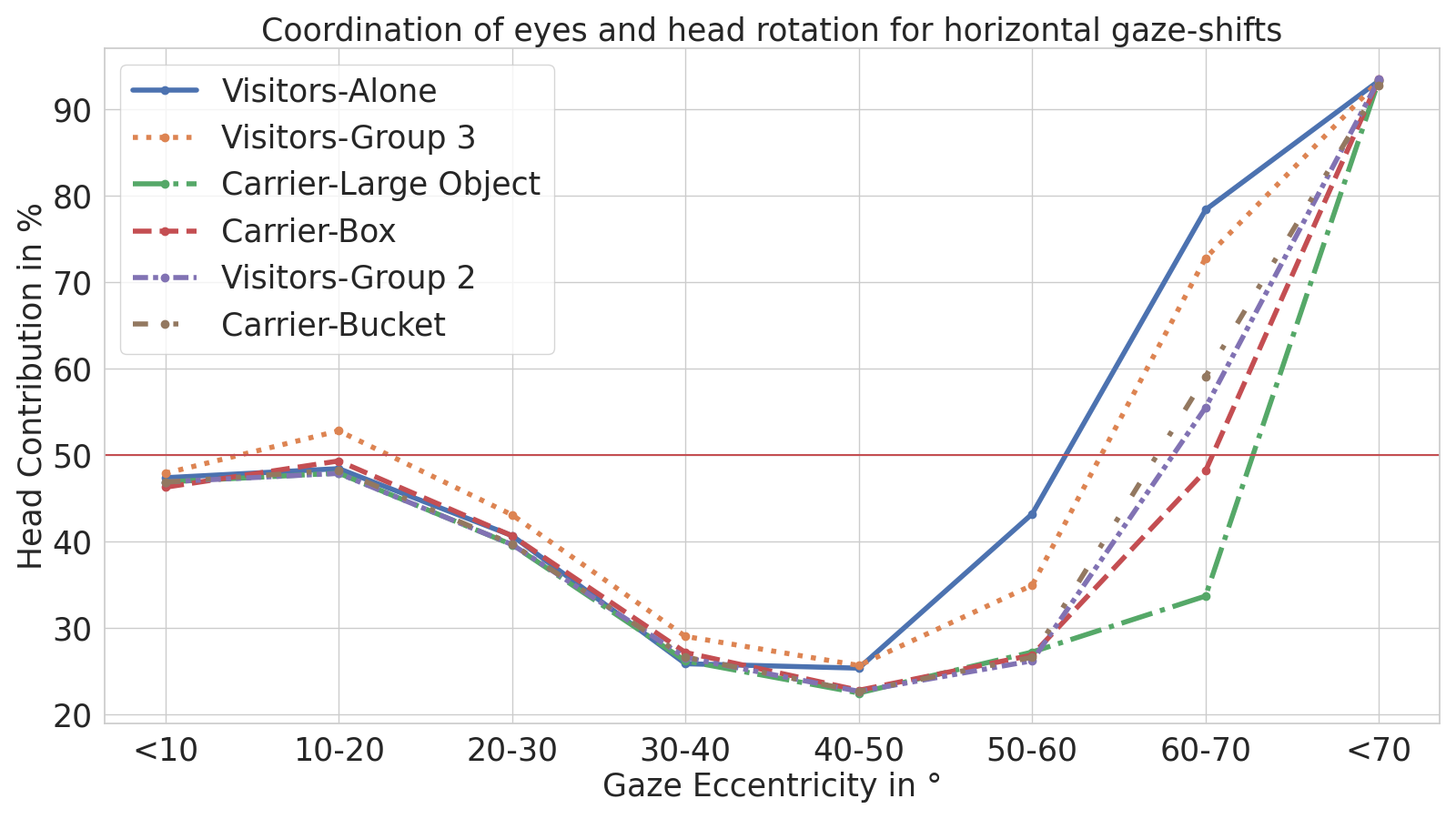}
    \caption{\textbf{Gaze eccentricity:} During motion peaks with eye and head rotations in the same direction, binned in 10-degree intervals. \textbf{Overall Head Contribution} to these shifts is shown on the other axis.}
    \label{fig:gaze-shifts}
    \vspace{2mm}
  \end{subfigure}
  \begin{subfigure}{\linewidth}
    \includegraphics[width=1\linewidth]{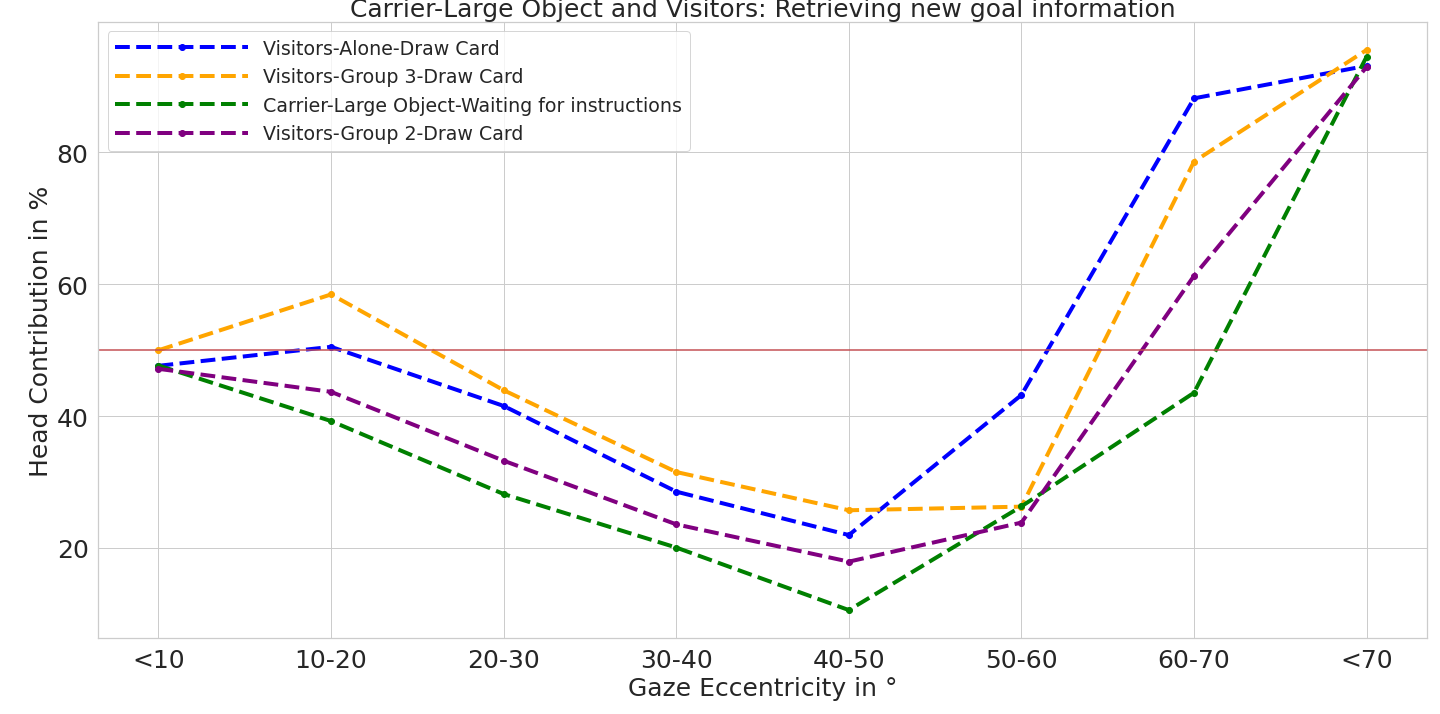}
    \caption{\textbf{Goal actions}: Visitors and Carriers-Large Object \textbf{drawing a card} or \textbf{waiting for instructions} for new goal points.}
    \label{fig:head_goal}
    \vspace{2mm}
  \end{subfigure}
  \begin{subfigure}{\linewidth}
    \includegraphics[width=1\linewidth]{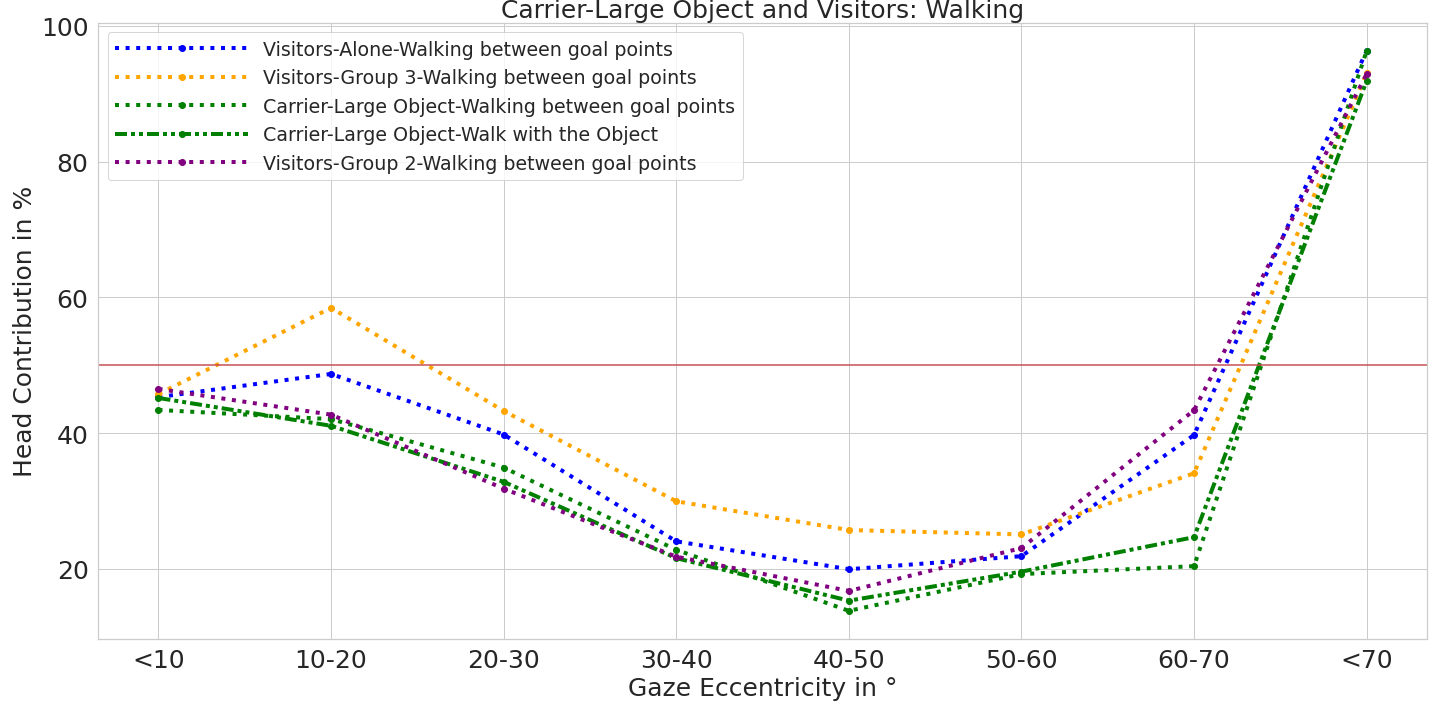}
    \caption{\textbf{Walking actions}: Visitors and Carriers-Large Object.}
    \label{fig:head_walk}
    \vspace{2mm}
  \end{subfigure}
  \begin{subfigure}{\linewidth}
    \includegraphics[width=1\linewidth]{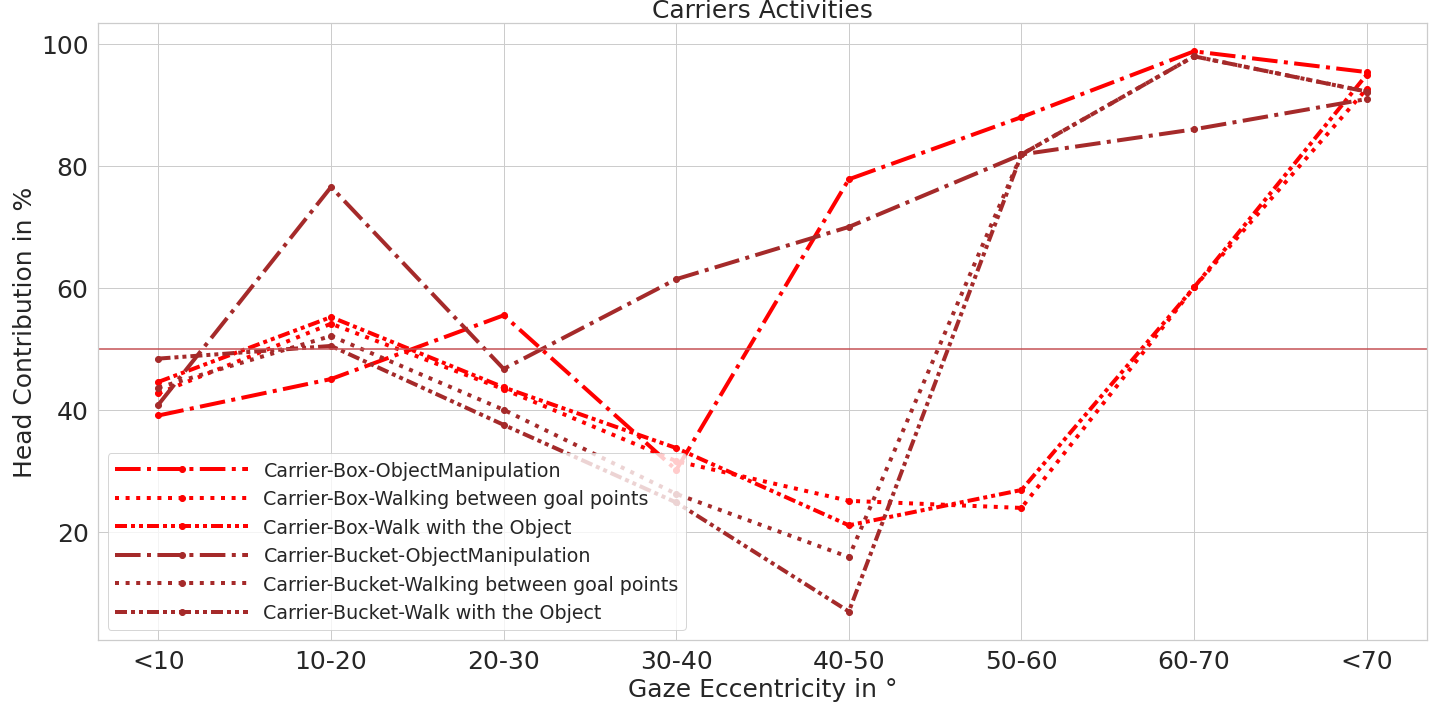}
    \caption{\textbf{Object manipulation and walking}: Carrier-Bucket and Carrier-Box, walking actions include with and without object.}
    \label{fig:head_carry}
  \end{subfigure}
  \caption{\label{fig:gaze-vs-head}Showing how much the head direction contributes to the total gaze direction, depending on the eccentricity of the gaze angle with respect to the body orientation. This figure applies to movements where the head and eyes move horizontally in the same direction.
  }
\end{figure}

To better understand eye-head coordination, we analyze instances where participants' eyes and heads rotated horizontally in unison. We focus on horizontal rotations because observers preferentially spread their gaze horizontally to explore their surroundings, and horizontal gaze movements are more common than vertical ones when walking over flat terrain \cite{franchak2021adapting}. To ensure the integrity of our data, we removed any anomalies, such as invalid eye-tracking records or rotations beyond the physiological limits (specifically, eye rotations exceeding $\pm55\degree$ and head rotations beyond $\pm80\degree$). We determined the horizontal eye rotation angle by multiplying the total width of the eye-trackers image (w) with the x-coordinate displacement (dx) and correlating it with the Tobii glasses' horizontal field of view (HFOV), represented as $\alpha_H = \text{w}\times\text{HFOV} / \text{dx}$. We assessed the horizontal head rotation (primarily captured by the motion capture system) angle using Euler angles from the rotation matrix $R$. Adopting the methodology of Franchak et al. \cite{franchak2021adapting}, we identified movement peaks separated by a minimum of 333 ms, to ensure we captured distinct shifts in attention. The 333 ms threshold matched the parameters used in Franchak et al.'s study to align our analysis with theirs and increase comparability. Results are organized into 10-degree categories to facilitate detailed analysis (see Figure \ref{fig:gaze-vs-head}).

Our systematic categorization of eye and head movement coordination across various activities revealed distinct patterns in horizontal gaze shifts. For minor attention shifts, contributions from both eyes and head were nearly equal, especially during locomotion (see Figure \ref{fig:head_walk}). For larger attention shifts, the relative contribution of the head decreases, reaching a minimum of around 45 degrees. This is slightly higher than the 35 degrees observed by Stahl et al. \cite{stahl1999amplitude} for eye movements in seated free viewing tasks. However, head movements were more dominant across all gaze shifts and micro-actions in groups of three visitors (see Figures \ref{fig:gaze-shifts} and \ref{fig:head_goal}). Beyond the 40-50 degree range, head movement contributions increased variably across different activities, with Visitors-Alone and those in Visitors-Group 3 adjusting their gaze more rapidly than those carrying objects. Eye movements dominated between 20 and 50 degrees, while more significant shifts above 70 degrees significantly increased head contributions. This pattern aligns with outdoor environment findings \cite{franchak2021adapting} but with a more pronounced reliance on eye movements in indoor settings. An exception to these trends is the micro-actions of Carrier-Box and Bucket during object manipulation, which strongly preferred head contributions to gaze shifts (see Figure \ref{fig:head_carry}).

\subsection{Correlating Motion with Gaze Alignment}
\label{subsec:motion}
To explore how human motion dynamics impact eye-head coordination, we propose several metrics of human motion, namely the straightness of participants' trajectories, walking speed, and acceleration. The straightness index (SI) ranges from 0 (non-linear) to 1 (perfectly straight), with higher values indicating more linear trajectories and lower values indicating more explorative trajectories. Mean walking speeds and accelerations are presented with standard deviations to highlight inter-individual variability. To calculate these motion metrics, we follow the preprocessing methods outlined by de Almeida et al. \cite{de_Almeida_2023_ICCV}. Results are outlined in Table \ref{tab:walking_speed_acceleration_straightness}.

Our analysis extends to correlating (Spearman correlation) these motion metrics with the alignment between head orientation and gaze vector, offering novel insights into human navigation strategies. We discovered a subtle yet statistically significant negative correlation between head and eye rotation alignment with walking speed ($\rho=-0.04$, $p < 0.01$), suggesting that increased linear velocity tends to enhance the synchronization of eye and head movements. A weaker yet also significant negative correlation with linear acceleration ($\rho=-0.01$, $p < 0.01$) indicates a less pronounced impact on the alignment of eyes and head compared to velocity.

\begin{table}[!]
\centering
\vspace{3pt}
\caption{Walking Speed, Acceleration, and Straightness Index (SI) for the roles in Scenarios 1--3 of the THÖR-MAGNI dataset}
\label{tab:walking_speed_acceleration_straightness}
\resizebox{\columnwidth}{!}{
\begin{tabular}{lcccc}
\toprule
\textbf{Role} & \textbf{Speed [m/s]} & \textbf{Acceleration [m/s²]} & \textbf{SI} \\
\midrule
Visitors-Alone & $0.88 \pm 0.55$ & $0.28 \pm 0.37$ & 0.6 \\
Visitors-Group 2 & $0.81 \pm 0.5$ & $0.24 \pm 0.31$ & 0.69 \\
Visitors-Group 3 & $0.80 \pm 0.5$ & $0.24 \pm 0.36$ & 0.77 \\
Carrier-Box & $1.07 \pm 0.47$ & $0.25 \pm 0.38$ & 0.95 \\
Carrier-Bucket & $1.16 \pm 0.4$ & $0.27 \pm 0.41$ & 0.97 \\
Carrier-Large Object & $0.65 \pm 0.51$ & $0.22 \pm 0.27$ & 0.75 \\
\bottomrule
\end{tabular}}
\end{table}

The motion metrics in Table \ref{tab:walking_speed_acceleration_straightness} vary across participant activities and reveal distinct movement patterns. ``Carriers'', characterized by high-velocity, linear movements and minimal head movement contribution to gaze shifts and centralized gazes, contrast with ``Visitors-Alone,'' who exhibit more dynamic movements with less linear trajectories and more explorative gaze distributions. ``Visitors-Group 2'' and ``Visitors-Group 3'' show similar speeds and accelerations to ``Visitors-Alone'' but follow straighter paths and demonstrate different gaze and head movement dynamics. ``Visitors-Group 3'' mainly displays a more considerable head contribution to gaze shifts.

Moving in pairs, the ``Carrier-Large Object'' participants displayed the slowest walking speeds and accelerations, a wide range of velocities, and the highest fixation rates, underlining a unique interaction pattern. Their contribution of head movement to gaze shifts notably stayed below 50\% for most activities despite their interactions with objects like the other carriers (see Figure \ref{fig:gaze-vs-head}), emphasizing the role's distinct coordination patterns. These findings emphasize the intricate relationship between physical motion and gaze behavior, contributing valuable perspectives to developing intuitive and responsive robot interactions in shared spaces.

\subsection{Quantifying Attention with Object Detection}
\label{subsec:objects}

\begin{figure*}[t]
    \centering
    \vspace{3pt}
    \includegraphics[width=1\linewidth]{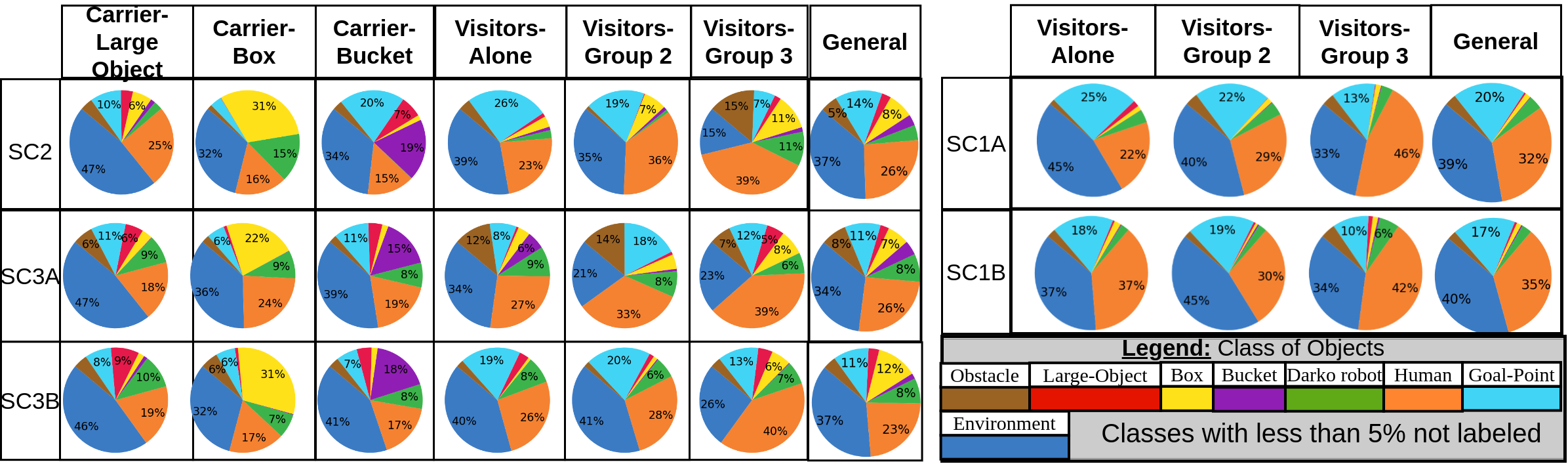}
    \caption{Distribution of fixations on objects by participants in Scenarios 2, 3A, and 3B (left) and 1A and 1B (right).}
    \label{fig:attention_distro}
\end{figure*}

The human gaze is heavily influenced by the static and dynamic objects in the environment. We employed an object detection method for the video frames from the eye-tracking glasses to achieve a finer decomposition of attention into classes of semantic objects. We used YOLOv8 \cite{yolov8_ultralytics}, pre-trained on the COCO dataset \cite{lin2014microsoft}, and refined with a custom dataset with labeled objects from THÖR-MAGNI. The classes, listed in Figure~\ref{fig:attention_distro}, include role-dependent objects (e.g. boxes and buckets), other walking people and the mobile robot DARKO,\footnote{\url{https://darko-project.eu/ }} (see also Figure \ref{fig:cover}).

Our custom dataset, consisting of 355 images annotated with seven classes, facilitated a focused analysis of participants' gaze during motion, particularly near the DARKO robot. Through this methodology, we observed notable shifts in attention distribution across different scenarios and activities, as visualized in Figure \ref{fig:attention_distro}. The pie charts illustrate a change in attention allocation from the environment and other participants in Scenario 1 to the more diversified attention towards the DARKO robot in subsequent scenarios, underscoring its significant presence in the shared space.

The statistical analysis, employing t-tests and calculating Cohen's d-effect sizes, supports these observations with significant findings. Specifically, the transition from Scenarios 1 and 2 to Scenarios 3A and 3B reveals a marked increase in attention towards DARKO, with effect sizes of $[-1.6, -0.8]$, all with $p < 0.1$, respectively, indicating a strong influence of the robot's presence on participant attention. This influence is further supported by the lack of statistically significant differences in attention between the different driving styles of DARKO in Scenarios 3A and 3B, suggesting that it is the robot's presence as a static or dynamic entity rather than its motion pattern that primarily captures human attention.


\section{Discussion}\label{sec:discuss}

Our analysis emphasizes the utility of using head orientation as a baseline assumption for gaze direction, which is particularly advantageous for onboard sensor-based gaze tracking. While this correlation is crucial for designing natural and seamless HRI systems, more than head orientation is required. It serves as an excellent standalone measure but must be complemented with gaze information to account for more dynamic settings, enabling refined interaction models that accommodate the complexity of human attention in diverse activities.

We observed that the contribution of eye gaze to attentional shifts decreases when objects are being carried or manipulated, indicating a preference for head movements over eye movements in these scenarios. Eye gaze is centralized during these tasks and focuses on objects and goal points relevant to the task fulfillment. For actions such as card drawing and navigating between goal points, where eye gaze plays a more significant role, understanding central gaze tendencies becomes essential to support gaze estimation via head orientation. Specifically, during navigation and visual exploration, gazes are primarily directed toward the upper hemisphere of the mobile eye-tracker (horizon) and on the lower hemisphere during card drawing or object manipulation. 

Moreover, our observations of participants in groups of three show that they divert their attention from robots to social interactions, displaying a slightly higher head contribution overall than other roles. This highlights the need for mobile robots to integrate sophisticated detection and anticipation algorithms in crowded areas, particularly around groups. Such capabilities are vital for navigating social environments, ensuring human safety, and optimizing robot operational efficiency. Understanding group dynamics provides valuable insights for designing robots that can navigate human social settings, adjust their behavior to minimize disruptions, and promote coexistence.

Additionally, our research on human-robot interaction has revealed a significant finding regarding the perception of robots. Specifically, the gaze distribution was similar for robots driving directionally and omnidirectionally, suggesting that the perception of these two mobility styles may not differ substantially. This similarity in gaze distribution indicates a potential versatility in human acceptance of different robotic mobility styles, which opens up avenues for innovative robot designs without compromising the user experience. The affirmation of technological advancements in robot locomotion encourages confidence in their acceptance within human-centered environments.

In conclusion, examining human gaze behavior in HRI contexts enriches our understanding of the interplay between human attention, perception, and robot design. These insights can advance the development of robotic systems that align with human behaviors and expectations, improving safety, efficiency, and integration into shared spaces. The broad applicability of these advancements, from collaborative manufacturing to autonomous vehicles, highlights the importance of gaze analysis in future HRI research and development.

\section{Conclusion}\label{sec:concl}

In this paper, we study the human gaze patterns during navigation and interaction with other people, mobile robots, and objects in the environment. We propose a set of tools to quantify human gaze dynamics and movement patterns and apply them to analyze the data in the THÖR-MAGNI dataset. In particular, we study how various activities and micro-actions affect the spread and bias of fixations. We investigate the coordination of head and eye rotation in shifting attention and identify optimal ranges for eye movements during medium shifts and head movements during large shifts. We propose several motion metrics to complement the gaze characteristics during dynamic interactions and finally use semantic object labeling to decompose the gaze distribution into activity-relevant regions.

Our study demonstrates that people's gaze during navigation is a powerful indicator of intention, attention, distraction, and the type of activity. Gaze analysis allows hypothesis verification of the efficacy of HRI methods and can be used to naturally assess navigation comfort around the moving robot. We believe that the analysis in this paper emphasizes the importance of gaze in HRI studies in dynamic environments and that the proposed methodology can be applied in other scenarios to better characterize and understand human behavior and transfer the findings to our robots.

\bibliographystyle{IEEEtran}
\bibliography{IEEEabrv,Roman_2024}

\end{document}